\documentclass[10pt,journal,compsoc]{IEEEtran}
\ifCLASSOPTIONcompsoc
  \usepackage[nocompress]{cite}
   \usepackage{cite}
   \usepackage{array}
  \usepackage{graphicx}
  \usepackage{arabtex}
  \usepackage{utf8}
  \usepackage{booktabs}
   \usepackage[dvipsnames]{xcolor}
  \usepackage{amsmath,amsfonts,amssymb}
  \usepackage{blindtext, rotating}
  \usepackage{multirow}
  \usepackage{algorithm}
  \newcommand{\subsubsubsection}[1]{\paragraph{#1}\mbox{}\\}
  \usepackage{color, colortbl}
  \usepackage{array,multirow,makecell}
\definecolor{Gray}{gray}{0.9}
\usepackage[noend]{algpseudocode}
\else

\fi
\ifCLASSINFOpdf
\else
\fi
\begin{document}
%
\title{ Neural Architecture based on Fuzzy Perceptual Representation For Online Multilingual Handwriting  Recognition \\ }
%
%
%
%

\author{Hanen~Akouaydi,~Student~Member,~IEEE,
        Sourour~Njah,~Member,~IEEE,
        Wael~Ouarda,~Member,~IEEE,
        Anis~Samet,~Professional ~tutor,
        Thameur~Dhieb,~Member,~IEEE,
        Mourad~Zaied,Senior~Member,~IEEE,
        and~Adel~M.~Alimi, Senior~Member,~IEEE
\IEEEcompsocitemizethanks{\IEEEcompsocthanksitem H.Akouaydi,  S.Njah, T.Dhieb, M.Zaied and A.M.Alimi, REGIM-Lab.: REsearch Groups in Intelligent Machines, University of Sfax, National Engineering School of Sfax (ENIS), BP 1173, Sfax, 3038, Tunisia.\protect\\

\IEEEcompsocthanksitem Anis Samet, 
Professional tutor, Sifast. 
E-mail: anis.samet@sifast.com}
\thanks{Manuscript received August 01, 2019;}}

%
%

\markboth{Journal of \LaTeX\ Class Files,~Vol.~, No.~, May~2019}%
{Shell \MakeLowercase{\textit{et al.}}: Bare Advanced Demo of IEEEtran.cls for IEEE Computer Society Journals}
%



\IEEEtitleabstractindextext{%
\begin{abstract}
Due to the omnipresence of mobile devices, online handwritten scripts have become the most important feeding input to smartphones and tablet devices. To increase online handwriting recognition performance,  deeper neural networks have extensively been used. In this context, our paper handles the problem of online handwritten script recognition based on extraction features system and deep approach system for sequences classification. Many solutions have  appeared in order to facilitate the recognition of handwriting. Accordingly, we used an  existent method and combined with new classifiers in order to get a flexible system.  Good results are achieved compared to online characters and words recognition system on Latin and Arabic scripts. The performance of our two proposed systems is assessed by using five databases. Indeed,  the  recognition rate exceeds 98$\%$.
\end{abstract}
}
\maketitle
\IEEEdisplaynontitleabstractindextext
%
\IEEEpeerreviewmaketitle

\ifCLASSOPTIONcompsoc
\IEEEraisesectionheading{\section{Introduction}\label{sec:introduction}}
\else
\section{Introduction}
\label{sec:introduction}
\fi

%
%
%
%
\IEEEPARstart{H}{andwriting} recognition system is the tool whereby a computer system can recognize characters and other symbols written by hand. It recognizes individual characters and an entire word, just as the human eye and mind would. Humans can understand any type of writing in different constraints. Thus, many researches are conducted to analyze handwriting and recognize it automatically.  In this context, we develop, in our work, two  systems for online handwritten script. In fact, the first system uses the perceptual theory, inspired from the human perceptual system, transforming language from graphical marks into symbolic representation. It proceeds by the detection of common properties, and then it gathers them according to some perceptual laws proximity, similarity...,etc.  Finally, it attempts to build a representation of the handwritten form based on the assumption that the perception of form is the identification of basic features that are arranged  till we identify an object. Therefore, the representation of handwriting is a combination of primitive strokes. Handwriting is a sequence of basic codes which are grouped together to define a character or a shape. This perceptual theory segments an handwritten script into elliptic strokes by Beta-elliptic model of handwriting generation. These strokes will be classified into primitives codes then we apply  LSTM classifier in order to recognize online handwritten script. We choose multilingual scripts : Latin and Arabic. \\
In the second system, we apply  convolutional LSTM on  (x, y, z) collected from Arabic/Latin datasets and also from data acquired  from Android application.\\
The structure of this paper is as follows: Section 2 presents an overview of systems for handwriting recognition in literature based on extraction features and deep learning approaches. Section 3 illustrates our two  systems for online handwritten script recognition. Section 4 reports the rate recognition performed on online handwriting and section 5 draws concluding remarks of future works. 
\begin{table*}
\caption{Summary of  systems for handwriting recognition.}
\begin{center}
\begin{tabular}{|c c c c c|}
\hline
\rowcolor{Gray}
Approaches & Authors  	& Database &	Method  &	Accuracy \\ 
\hline
     \multicolumn{1}{c}{\multirow{2}{*}{Handcrafted Approaches}} & \cite{Najiba2017}	& LMCA +
manually Segmented characters form ADAB  &	 DNN  & 82.12$\%$\\ 
    \multicolumn{1}{c}{}& \cite{Tagougui14} & LMCA +
manually Segmented characters form ADAB & MLP+HMM system & 96.45$\%$\\
   \multicolumn{1}{c}{}& \cite{keysers2017} & UNIPEN+  Google cloud  &	Language Model & 99.02$\%$ \\ 
   \multicolumn{1}{c}{}& \cite{Ghods2016} & Farsi & NN+GA	& 94.20$\%$\\
   \multicolumn{1}{c}{}&  \cite{Hamdi2017} & ADAB &  Grapheme segmentation + MLP	& 93.50$\%$ \\
\hline
\hline
 \multicolumn{1}{c}{\multirow{2}{*}{Deep Learning Approaches}} & \cite{Yang2015} &	Chinese & CNN	&   97.20$\%$  \\ 
    \multicolumn{1}{c}{}& \ \cite{Zhang2017}  &	Mathematical Expressions &  GRU	& 52.43$\%$ \\
   \multicolumn{1}{c}{}&   \cite{Xiao2017}  &	Chinese &  CNN+Drop Weight	& 96.88$\%$ \\
   \multicolumn{1}{c}{}&  \cite{XZhang}  &	Drawing $\&$ Recognition Chinese characters  & RNN  & 93.98$\%$  \\
\hline  
\hline
\end{tabular}
\end{center}
\end{table*}

\section{RELATED WORK}
Handwriting recognition is  transforming graphical marks into symbolic representation. Although humans are similar and have a common education, they produce significantly different handwritten styles. The factors of the variability are related to the writer, emotional factors, language of writing, the writing with different size, styles and speed...etc.
We found two main approaches for Handwriting Recognition : The first approach is  based on segmentation  and classification (or segment-and-decode) and the second is related to time-sequence interpretation.
\subsection{Handcrafted Approaches}
Those approaches segment then decode the script in order to  identify it. Traditional classification are based on extraction features that are engineered and extracted from kinetic signals.\\
Google online handwriting system is one of the most powerful systems that presently supports 22 scripts and 97 languages. The same architecture of its system is used for mobile devices with more limited computational power by means of changing some of the settings of the system \cite{keysers2017}. We also find another approach for online Arabic handwriting recognition, based on neural networks approach. This work suggests  a method based on TDNN classifier and multi-layer perceptron,it achieved 98.50\% for characters, 96.90\% for words \cite{Ghods2016}. There is another work which recognizes mathematical expression by using HMM and the system was conducted by covering two things: feature modification experiment and code words number experiments. The best result is gained  with respect to four combination features \cite{Pranoto2016}. We also notice a system for  online Farsi characters recognition. The proposed system is tested on TMU database and 94.2\% obtained as recognition rate for the recognition of the group and the character, respectively \cite{Ghods2016}. Hamdi et al. \cite{Hamdi2017} suggest  a hybridization of neural net-works and genetic algorithm for online Arabic handwriting recognition. In fact, based on Beta-elliptic model and baseline detection. The proposed system reached 93.5\% as average of recognition. Another work deals with Online Arabic Handwritten Character Recognition based on a Deep Neural Networks DNN in which we attempt to optimize the training process by a smooth construct of the deep architecture.  This system uses LMCA database for training and testing data. Tagougui et al. gained 96.16\% as testing rate \cite{Najiba2017}.
There is a new work which proposes a model based on Bézier curve interpolation and recurrent neural networks for online handwriting recognition and that able to support 102 languages\cite{Victor2019}. Besides,  another work describes a hybrid model based on bi-directional LSTM and Connectionist Temporal Classifier (CTC) for handwriting recognition on mobile devices.\cite{article2019}\\
The success of deep neural networks approaches has been achieved based on handwriting recognition. However, recognition is not only a way to understand a language, but also a challenging task to teach machine how to write automatically. Most research works did not propose universal methods to recognize special languages as Farsi, Chinese or Mathematical Expressions\cite{Tagougui2013}.
The major  motivation of this paper consists in dealing with online script by using some novel techniques in training deep recurrent neural networks. In fact,our proposed models  for handwriting recognition applied various lexicons by using perceptual codes, LSTM and convolutional LSTM.
\subsection{Deep Learning Approaches}
Generative approaches based on Convolutional Neural Networks (CNN)  and LSTM to recognize handwritten script. The extraction features process requires a deep knowledge of application domain, or human experience or hardware memory to store script in order to identify script from user. Accordingly, automatic and deep methods are attracting in the field of handwriting Recognition. By adoption data-driven approaches for signal (x, y, z) script classification, the process of selecting meaningful features from the data is deferred to the learning model. CNNs are capable of detecting  both spatial and temporal dependencies among signals, and can effectively model scale invariant features.
It is noteworthy that two main advantages can be considered,  when applying deep neural networks  to handwriting recognition which are as follows:
\begin{itemize}
\item [$\bullet$]Local Dependency:New Deep Neural Networks capture local dependencies of handwriting signals. In image recognition tasks, the nearby pixels typically have strong relationship with each other. Similarly, in handwritten script(online (x, y, z)) given an script (signal) the nearby acceleration readings are likely to be correlated.
\item [$\bullet$]Scale Invariance: Deep neural Networks  preserve feature scale invariant. In image recognition, the training images could probably different scales. In handwritten script , a person may write  with different styles  (e.g., with different motion intensity).
\end{itemize}
 Those  approaches used are:  hidden Markov models (HMMs)\cite{ttt10}, time-delay neural networks (TDNN)\cite{ttt7}, and recurrent neural networks (RNN)\cite{ttt8}, of which long-short-term-memory
networks (LSTM)\cite{ttt9} are a specific case,  presently receiving significant attention with regards to machine learning.
\section{PROPOSED SYSTEM}
\begin{figure*}[!ht]
\centering
   \includegraphics[width=\textwidth]{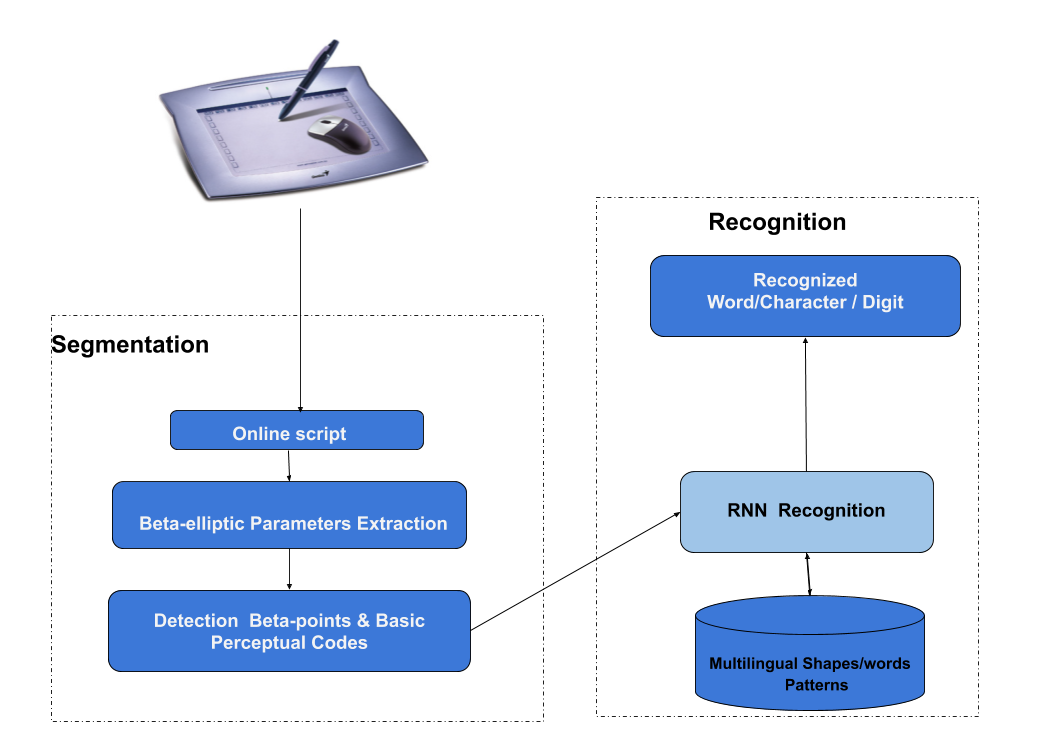} 
  \caption{Architecture OnHS-LSTM.}
 \end{figure*}
 \begin{table*}[h]
\centering
\caption{Basic Perceptual Codes}
\label{table_example}
\begin{tabular}{|c|c|c|c|}
\hline
\rowcolor{Gray}
\textbf{Number} & \textbf{Basic Perceptual Codes} & \textbf{Abbreviation} & \textbf{Shape} \\
\hline
\hline
1 & Valley  & V & \parbox[c]{1em}{ \includegraphics[width=0.25in]{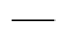}}\\
\hline
2 & Left-oblique-shaft & L-O-S & \parbox[c]{1em}{ \includegraphics[width=0.25in]{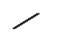}}\\
\hline
3 & Shaft & S & \parbox[c]{1em}{ \includegraphics[width=0.25in]{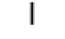}}\\
\hline
4 & Right-oblique-shaft & R-O-S & \parbox[c]{1em}{ \includegraphics[width=0.25in]{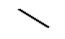}}\\
\hline
5 &  Right-Half-Occlusion &  R-H-O & \parbox[c]{1em}{ \includegraphics[width=0.25in]{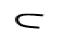}}\\
\hline
6 & Left-Half-Occlusion & L-H-O & \parbox[c]{1em}{ \includegraphics[width=0.25in]{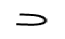}}\\
\hline
7 & Up-Half-Occlusion & U-H-O & \parbox[c]{1em}{ \includegraphics[width=0.25in]{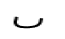}}\\
\hline
8 & Down-Half-Occlusion & D-H-O &\parbox[c]{1em}{ \includegraphics[width=0.25in]{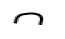}}\\
\hline
9 & Occlusion & Occ & \parbox[c]{1em}{ \includegraphics[width=0.25in]{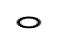}} \\
\hline
\end{tabular}
\end{table*}
Online handwritten scripts are always stored as form of sequence points. Divers features can be extracted from them such as stroke coordinates, times stamp, writing angle and curvature, etc. The aim of  the turn-out of such system is to recognize the sequence points.
In this paper, two models are suggested, both  of them are based on deep neural networks first  corresponds to Online Handwriting Segmentation $\&$ Recognition using LSTM,
\textbf{OnHSR-LSTM}, which  requires a step of segmentation,  and the second, consists in Online Handwriting Recognition using convLSTM, \textbf{OnHR-convLSTM}.
The two models will be described below:
\subsection{ Online Handwriting Segmentation $\&$  Recognition using LSTM, \textbf{OnHSR-LSTM}}
\begin{figure}
      \centering
      \includegraphics[height=4cm,width=8cm]{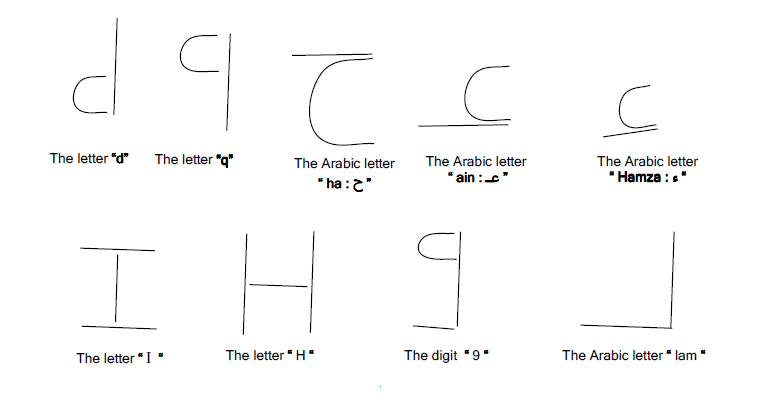}
      \caption{Illustrated example of composed forms.}
\end{figure}
Our first model OnHSR-LSTM, is based on the fact that handwriting is a form created by hand movement and all we perceive is only a form \cite{Njah2010,Njah2012,Njah2011}
It proposes that handwriting is a concatenation of perceptual codes grouped together, so as to form a character, a digit or a word. Each online script can be composed of elementary basic shapes.
With  (x, y, z) coordinates of online trace and the beta-elliptic model we detect basic shapes forming the script. Then,  we classify them relying on LSTM in order to recognize the character or the written digit. Figure2 represents the architecture of LSTM-Character-recognizer, which recognizes Online Handwritten letters.
Our first model is endowed with crucial phases:the phase of segmentation and the  phase of classification and recognition.
 \subsubsection{Segmentation Step}
 The segmentation step is concerned with the regeneration of the online handwritten script by means of strokes or by a set of control points that describe the handwritten trace.
\subsubsubsection{Preprocessing}
Preprocessing is a key  step with respect to recognition in order to obtain better recognition rate.   The principles of preprocessing steps are used  to eliminate trembles in writing , to reduce noise and to remove the hardware imperfections. 
The  preprocessing operations used in our system are:
\begin{itemize}
\item [$\bullet$]  \textbf{Interpolation}: adding missing points caused by the variation of writing velocity.
\item [$\bullet$]  \textbf{De-hooking}: eliminating trembles in writing due to inaccuracies in rapid pen-down/up detection and trembles in writing;
\item [$\bullet$]\textbf{Removing noise}: we apply a Chebyshev low pass filer to eliminate the noise generated by spatial and temporal  sampling;
\end{itemize}
\subsubsubsection{Beta-elliptic Features}
Handwriting is a message which can be decoded into basic shapes. The main assumption of our proposed model consists in the fact that handwriting is a concatenation of visual codes grouped together so as to get a shape, a character or a digit. The principal basic shapes to write are: ${-}$, ${/}$, ${|}$,\textbackslash.\cite{pinale,mal1993}.
The basic shape which can compose any script. Indeed, the basic shapes are illustrated in Table 2.  
The analysis of any language has demonstrated that a script is an arrangement of vertical and diagonal lines as shown  Figure 1, which demonstrates that a letter is a combination of two or more basic shapes.
In fact we have 9 basic perceptual shapes that can be simple or complex. These basic shapes have been identified as Global Perceptual Codes with GPC "Ain" which only characterize the Arabic script.\cite{Njah2010,Njah2012,Hanen2017,Ltaief2012,Sourour2008}
Thus, in this paper we use only 9 basic perceptual shapes which can form any script.
In Figure 1, we notice that handwriting
is an arrangement of perceptual codes in a special
order, and we can from different scripts with the same group of basic shape such as the two Arabic letter  \setcode{utf8}\<عـ> hamza \setcode{utf8}\<ء>.
In this step, we extract the basic shape that gather together to identify  a script. Hence, we use the beta-elliptic model as a model of segmentation.
Basic shape are not directly detectable. Therefore, we use in the step of segmentation the beta-elliptic model.
\\
Beta-elliptic model propose that handwriting  can be reconstructed by the sum of impulse signals by the curvilinear velocity and can be also approximated by a sequence of elliptic shapes.\cite{Hala2002, Ltaief2016,Sourour2008}  
So, handwriting can be segmented into simple movements called strokes, which are the result of superposition of time overlapped velocity profiles. Consequently, each stroke with  its curvilinear velocity obeys the Beta-elliptic approach and  online script can be segmented into different ones by tracting Beta-elliptic parameters.\cite{Ltaief2012,Njah2011,Njah2012,Njah44}
\\
\begin{figure*}[!ht]
\centering
   \includegraphics[width=\textwidth]{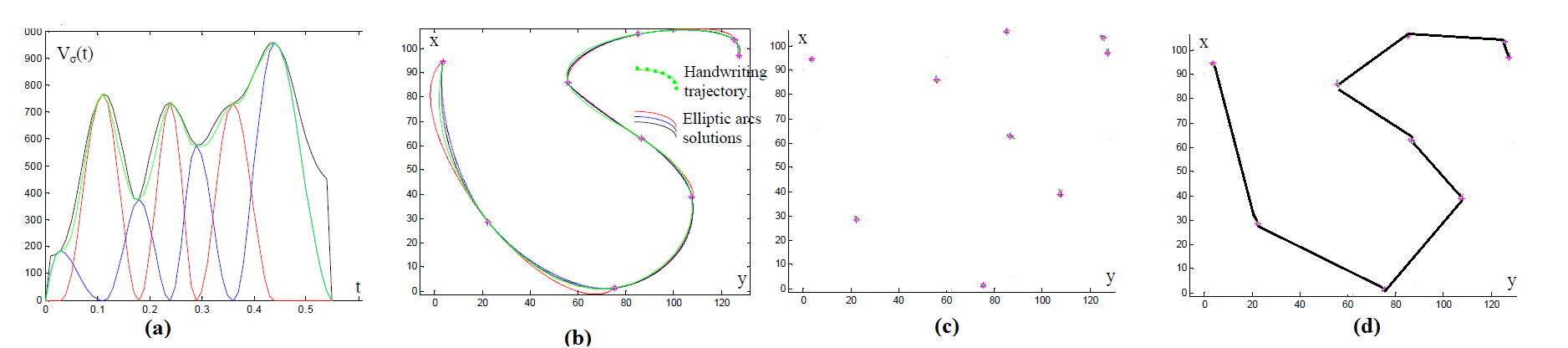}
  \caption{Segmentation Step of Arabic Letter "yaa".}
  
 \end{figure*}
 \begin{figure*}[!ht]
\centering
   \includegraphics[width=\textwidth]{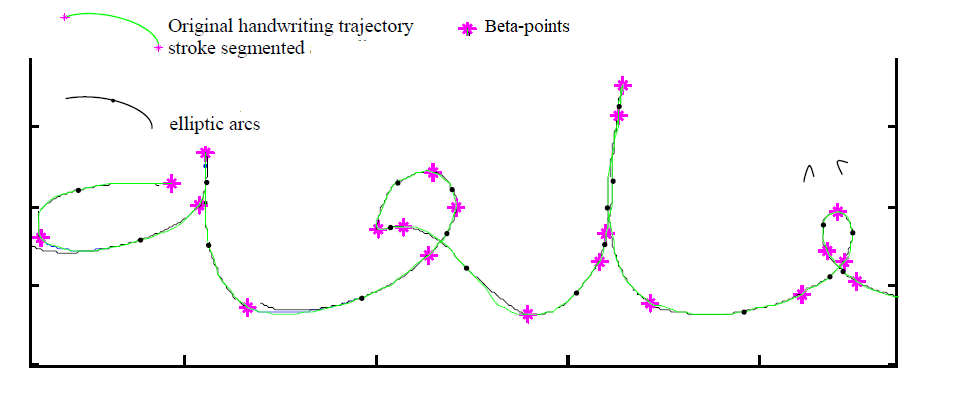} 
  \caption{Segmentation and  Detection of Beta-points of Arabic word "Kalaa" .}
 \end{figure*}
 \begin{figure*}[!ht]
      \centering
      \includegraphics[width=\textwidth]{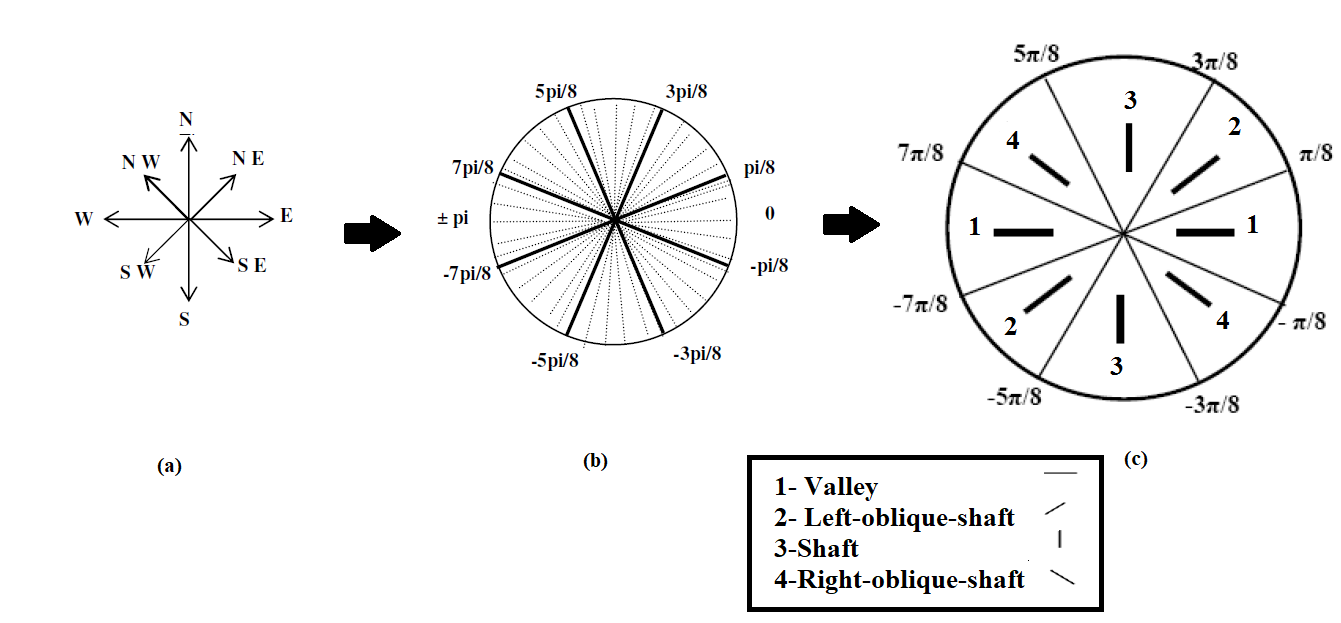}
  \caption{Directions: (a): Freeman$'s$ chain code, (b): the proposed regions, (c):Deviation angles regions and EPCs on the trigonometric circle,(d):Elementary Perceptual Codes}
 \end{figure*}
The Beta function is presented in the equation(1)
\begin{equation}
   \begin{split}
 \beta(t,p,q,{t_0},{t_1}) & = {(\frac{t-{t_0}}{{t_c}-{t_1}})^p}{(\frac{{t_1}-t}{{t_1}-{t_c}})^q},{IF     {t\in]{t_0},{t_1}[}}   \\
                          & =0 ,      IF not 
  \end{split}
\end{equation}
where  $\boldmath{t_0}$: starting time of Beta function.\\
${t_c}$: is the instant when the curvilinear  velocity reaches the amplitude of the inflexion point.\\
${t_1}$: the ending time of Beta function.\\
$ With             (p,q,{t_0}<{t_1})\in {R},and $\\
\[{t_c}=\frac{p*{t_1}+q*{t_0}}{p+q},p=q*\frac{t-{t_0}}{{t_c}-{t_1}}\] \\
p , q intermediate parameters which have an influence on the symmetry and the width of Beta shape.\\
As a result, a stroke is characterized by 10 parameters.The following beta-elliptic parameters a, b, ${x_0}$,${y_0}$ and $\theta$ reflect the geometric properties of the set of muscles and joints used in a particular handwriting movement and  describe the static aspect of handwriting. 
These  parameters are:\begin{itemize}
\item [$\bullet$]${x_0}$,${x_1}$:the  coordinates of the ellipse center,
\item [$\bullet$]a: big axe of the ellipse
\item [$\bullet$]b: small axe of the ellipse
\item [$\bullet$]$\theta$: the angle of the deviation of the elliptic arc and the horizontal which is obtained by the following equation 2:
\end{itemize}
\begin{equation}
 \theta = \arctan(\frac{{Y_1}-{Y_0}}{{X_1}-{X_0}})
\end{equation}

The Beta elliptic was performed in systems for writer identification and verification.\cite{Thameur2015,Thameur2016,Thameur2018}
An online handwriting script can be segmented into segments and each segment is a group of strokes. 
In fact, the number of strokes is identified automatically from the curvilinear velocity representation. In the trajectory domain, each stroke is limited by both M1 and M3 that correspond to the maximum curvature and minimum curvilinear velocity. M2 is defined by the maximum tangential velocity. 
\subsubsubsection{Elementary Perceptual Codes (EPC) Detection with Beta-points}
The main idea of this part is to associate  an EPC (Elementary Perceptual Code) to each stroke. It means that we classify strokes of the handwritten script generated by the beta-elliptic model into EPCs.This classification is performed by using the parameter, corresponding to the deviation angle of each ellipse "$\theta$", by the horizontal axis. 
An EPC is limited by beta control points M1, M3 and goes through H which the orthogonal projection of M2. The number of strokes is identified automatically by beta-elliptic model. This number is equal to the number of EPCs.
The EPCs are similar to Freeman's code, which are frequently used for online handwriting recognition, being able to express stroke directions. Based on this technique, features are extracted from each online trajectory segment.
By means of analogy with Freeman’s codes, an EPC is a whole region incorporating a group of directions as depicted in Figure 2.(a) showing the 8 directions of Freeman's codes and  2.(b) illustrating  all EPCs in their corresponding parts.\cite{Njah2012}\\
Different factors(style, speed, position, size, orientation ) can affect the handwritten script, thus making its treatment and analysis difficult. Figure 6 presents some regions of trigonometric circle, we consider an overlapped interval value noted cst equal to /16 where there is a problem of indecision pertaining to the corresponding EPC. This indicates an example of the perceptual problem decision related to the belongings of elliptic stroke.
To prevent some perception problem, we associate an EPC to each stroke with a certain membership degree. Thus, we opt for employing fuzzy logic representation to resolve the problems of fuzzy perception and the uncertainty of the assessed EPC \cite{Njah2010,Njah2012,Hanen2017}. Correspondingly, for each stroke a certain belongingness degree will be allowed. 
\begin{figure}
\centering
   \includegraphics[height=8cm,width=9cm]{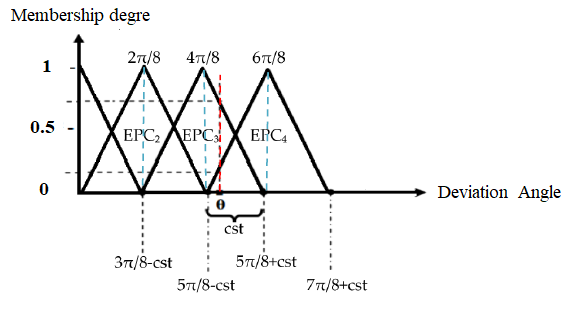}
  \caption{Example of perceptual problem decision using fuzzy logic.}
 \end{figure}

\subsubsection{LSTM FOR CLASSIFICATION}
After extracting the Basic Perceptual Codes (BPC) of online script, a classifier is required to identify it. 
We opt to use LSTM, Long Short-Term Memory as classifier. LSTM is specially constructed RNNs, Reccurent Neural Networks, node to preserve long lasting dependencies by avoiding the vanishing gradient problem. It is explicitly conceived to prevent the long-term dependency problem, a tool of sequence classification in order to identify time-sequences.  In fact, their default behavior of LSTM cells lies mainly in remembering information forlong periods of time. It provides a memory of the previous network internal state. Long Short-Term memory (LSTM) layer: The LSTM network nodes posses a specific architecture, referred to as memory block. Each memory block incorporates a memory cell, and its interaction with the rest of the network is controlled by three gates, i.e: an input gate, an output gate and a forget gate. This permits the memory cell to preserve its state for a long period.
The following illustrates the architecture of LSTM
\begin{figure}[thpb]
      \centering
      \includegraphics[height=4cm,width=8cm]{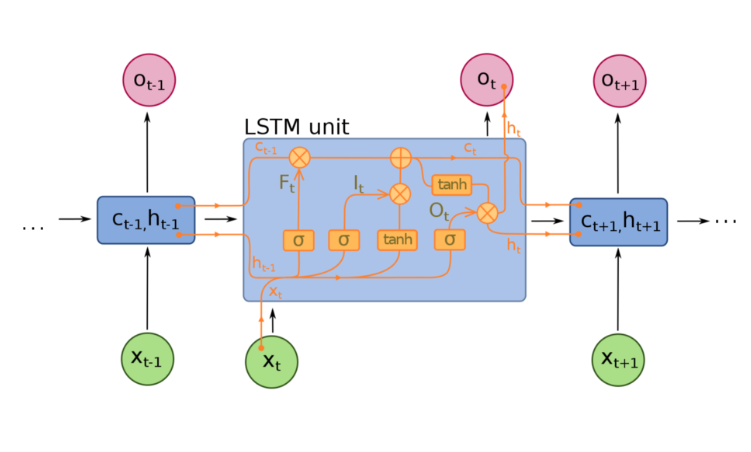}
      \caption {Architecture of LSTM cells.}
\end{figure}
To identify the online script or to identify basic perceptual codes, we use LSTM. The handwritten script is segmented into  strokes limited by beta-points. We apply fuzzy logic in order to obtain the membership of one stroke to the 4  EPCs. A segment is a group of strokes, and stroke is composed of 4 EPCs with degree of membership. Each stroke is classified into 4  EPCs, and a segment is a group of n strokes. A BPC is a group of n segments or a script(character, digit ,word) is a group of n EPCs. 
Handwriting is a group of perceptual  codes which are the bases pattern to read, write, identify, segment and recognize a script. In this section, we detail how from these perceptual codes, we can form a character, a shape or a digit.
Handwriting has an equation as our perceptual theory which is:
\begin{equation}
\label{h1}
    handwriting=\{{Segment_1i},{Segment_2i},   {Segment_ni}\}
\end{equation}
 n:the total number of \textbf{segment}s composing handwriting. 
 \begin{equation}
\label{h1}
  handwriting=\{{BPC_1i}
   ,{BPC_2i},...,{BCP_ni}\}
\end{equation}
\\
 i:{1,9} correspond to \textbf{BPC}s. (see Table II)
\begin{equation}
\label{h2}
BPC=\{ {EPC_1j},{EPC_2j},...,{EPC_pj} \}
\end{equation}
\begin{equation}
\label{h2}
BPC=\{ {Stroke_1j},{Stroke_2j},...,{Stroke_pj} \}
\end{equation}
p:Number of \textbf{stroke}s identified in the trace.
 \\
 j:{1,2,3,4} correspond to \textbf{EPC}s. (see Table I)
With an adequate arrangement of elementary perceptual codes (EPCs),
\begin{equation}
\label{h2}
Stroke/EPC=beta-point1+beta-point2+beta-point3
\end{equation}
The  beta-points are three ,anchorage perceptual points\cite{Njah2010,Njah2012,Sourour2008},forming a stroke. The stroke goes from M1 to M2, which  are the maximum curvilinear velocity and  the minimum curvilinear velocity, respectively. It passes through H which is the orthogonal projection of M2, standing for the maximum tangential velocity or the inflexion point.
These Basic perceptual Codes (BPCs)are divided into two classes:
\begin{itemize}
\item [$\bullet$]A simple class contains: Valley, Left oblique shaft, Shaft, Right oblique shaft.
 \item [$\bullet$]A complex class is composed of: Right half occlusion, Left half occlusion, Up
half occlusion, Down half occlusion, Occlusion.
\end{itemize}
 As already mentioned, handwriting is a sequence of basic perceptual codes and by the combination of simple ones, we obtain global ones. Numerous possible combination of EPCs can form a BPC. In fact, there is a large number of combinations that forms a BPC. Therefore, a large set of BPCs features is assumed to be available in order to recognize the corresponding BPCs of the initial script. As mentioned above, the Basic Perceptual Codes are a set of sequence of Elementary perceptual Codes. The same BPC can be obtained by different sets of EPCs.
The BPC "Valley" can be obtained by a minimum set of three successive EPCs, corresponding to Valley and at a  maximum set of six Valley EPCs.
Those difficulties which are the sequences of EPCs, can vary in length and be comprised of a very large combination of one BPC, and may require a model to learn the long-term context or dependencies between BPCs in the input sequence of initial script.  
Thus,  the  problem  of  the  suitable   choice  of corresponding BPC can  be  consequently regarded  as related to sequence classification. Sequence classification is a predictive modeling problem, there is some sequence of inputs  over  space  or  time  and  the  task consist in predicting  a category for the sequence. 
\begin{figure}[thpb]
      \centering
      \includegraphics[height=4cm,width=8cm]{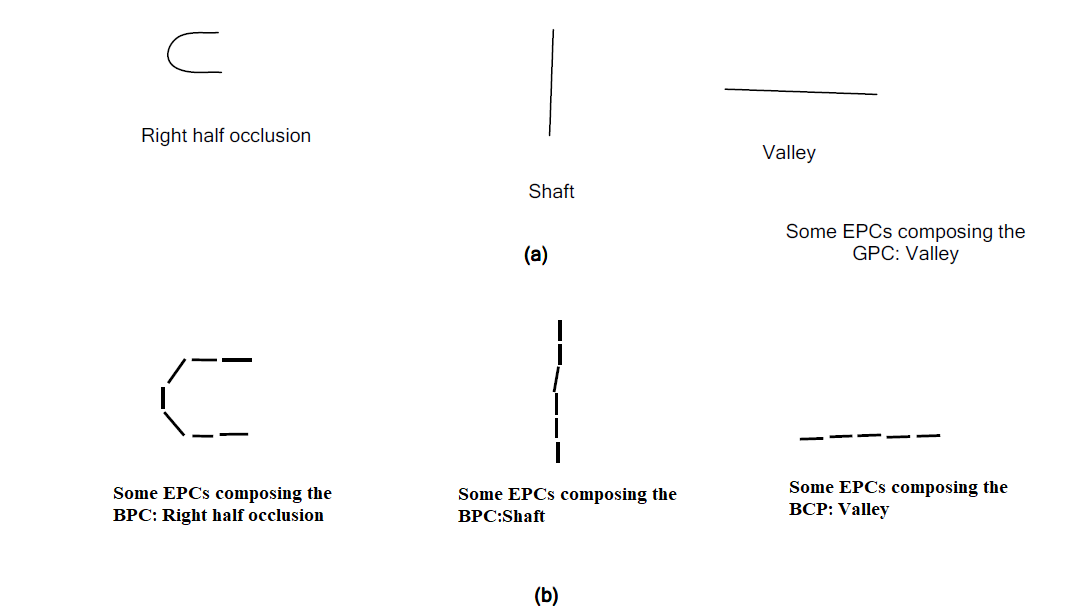}
      \caption {Some samples of BPC's combination.}
\end{figure}
To improve the performance of our network, we use the CTC as output layer after softmax layer in system 1 which pertains to Perceptual Codes-LSTM for Online handwriting recognition. In fact, it is a layer of labelling.
The CTC algorithm employs a many-to-one sequence-to-sequence mapping function that converts the sequence of labeling with timing information  into the shorter sequence of labels, by means of removing timing and alignment information \cite{Graves26}.  
Indeed, the main point is to introduce the additional CTC blank label for the sequence that has timing information, then remove the blank labels and merge repeating labels do as to obtain the unique corresponding sequence.
Based on our OnHSR-LSTM system,the output of the final LSTM layer is passed through a softmax layer that yields the predictable characters for each time step.
The output of the LSTM layers at each time step is located into a softmax layer in order to obtain a probability distribution words, digits or characters, as well as a blank label added by CTC.
\subsection{Convolutional LSTM for Handwriting Recognition} 
  \begin{figure*}[!ht]
      \centering
      \includegraphics[width=\textwidth]{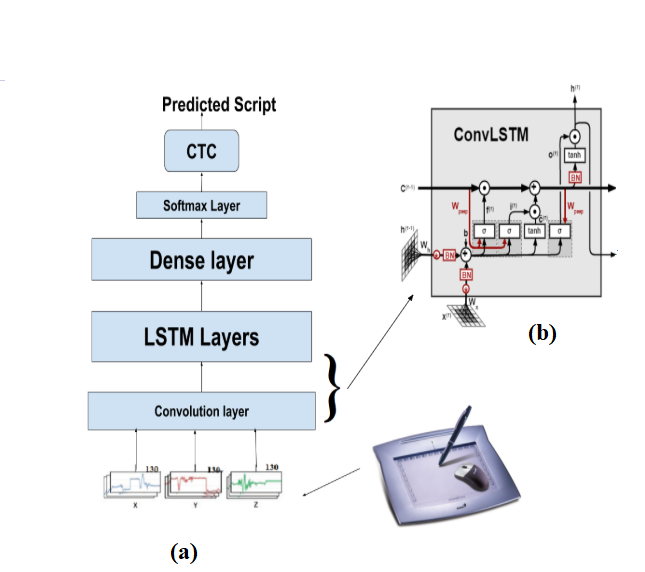}
      \caption{(a)Architecture of OnHR-convLSTM, (b) convLSTM cell.}
\end{figure*}
Our second proposed system  OnHR-convLSTM, is presented in the Figure 9 with the architecture of convLSTM cell. It is a generative model that has as input the online signal (x, y, z) of the script and predicts characters as well as words from trained model.
This generative system excels with sequence learning tasks and can align unsegmented data with their corresponding ground truth.
LSTM network models are a type of recurrent neural network capable of learning and remembering over long sequences of input data.  In fact, they may be convenient for such
problem. 
The model can support multiple parallel sequences of input data, such as each axis of the (x, y, z) data. The model learns to extract features from sequences of observations and how to map the internal features to various types of activity.
The merit of using LSTMs for sequence classification lies in the fact that they can learn from the raw time series data, directly, and in turn, do not necessitate domain expertise to manually engineer input features. The model can learn an internal representation of the time series data, thus ideally achieving comparable performance to models fit on a version of the data-set with engineered features.
Convolutional-LSTM model is a different approach, wherein the image passes through the convolution layers and the result consists in a set flattened to a 1D array with the obtained features.  Repeating this process to all images in the time set, give rise to a set of features over time, which is, in fact, the LSTM layer input.
The ConvLSTM layer output is a combination of a Convolution and  LSTM output.
The other ConvLSTM parameters  are derived from the Convolution and the LSTM layer. The 
The Convolutional LSTM architecture includes using Convolutional Neural Network (CNN) layers for feature extraction on input data combined with LSTMs, in order to support sequence prediction.
It was developed for visual time series prediction problems and the application of generating textual descriptions from sequences of images (e.g. videos). 
A Convolutional LSTM can be defined by adding CNN layers on the front end followed by LSTM layers with a Dense layer on the output. This architecture is considered helpful by defining two sub-models: the CNN Model for feature extraction and the LSTM Model for interpreting the features across time steps. 
Unlike an LSTM that reads the data directly in order to calculate internal state and state transitions, and unlike the CNN LSTM that is interpreting the output from CNN models, the ConvLSTM uses convolutions directly as part of reading input into the LSTM units themselves.
Both Convolutional Neural Networks (CNNs) and Long Short-Term Memory (LSTM) have shown improvements over Deep Neural Networks (DNNs) across a wide range of speech recognition tasks.
CNNs, LSTMs and DNNs are complementary with respect to their modeling capabilities. Indeed, CNNs are efficient at reducing frequency variations, LSTMs are good at temporal modeling, and CNNs are appropriate for mapping features to a more separable space. Therefore, we combine them into one unified architecture.
\section{Regularization and Data Augmentation}
To improve the performance of our system, we  used some techniques, as illustrated above.
\subsection{Dropout}
The main goal when using dropout is to regularize the neural network that, we train. The technique involves of dropping neurons randomly with some probability p. Those random modifications of the network’s structure are believed to avoid co-adaptation of neurons by making it impossible for two subsequent neurons to rely solely on each other. In fact, dropout is implicitly bagging, at test time, a large number of neural networks that share parameters. Besides, dropout is used it to prevent  the over-fitting of our network from  and to upgrade the latter's performance, dropout is applied. This technique consists in temporarily removing a unit from the network. The selection of such removed unit is randomly occurring only during the training stage.
\subsection{Data Augmentation}
The use of deformation technology, in the context of deep neural networks, aims  chiefly providing shape variation and generating  numerous online training data. Deep LSTM extends the data-set by applying affine transformations including scaling, rotations, and translations. Besides, stroke jiggling is also used to generate local  distortions to enrich our  data with local diversity.
\begin{figure}[thpb]
      \centering
      \includegraphics[height=4cm,width=8cm]{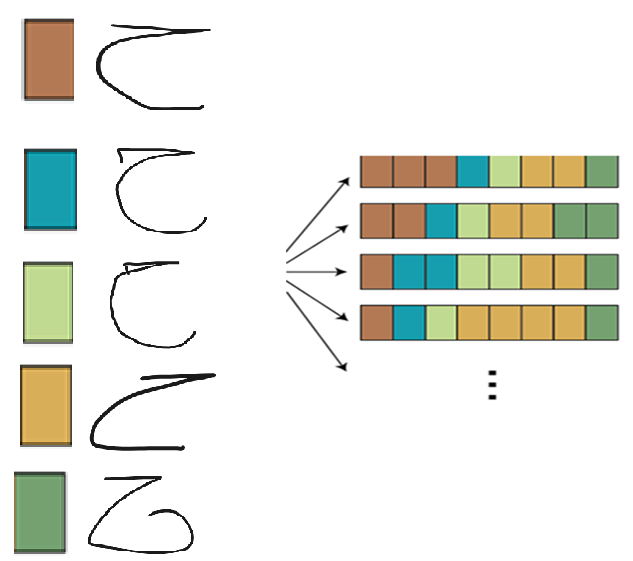}
      \caption{Data Augmentation of Arabic letter "haa" \setcode{utf8}\<ح>  .}      
\end{figure}  
Data augmentation is added in order to prevent over-fitting, and to make the learning more uniform. The neural network training is at the risk of over-fitting and hurting the classification robustness. One way to deal with this problem is data augmentation where the training set is artificially augmented by adding replicas of the training samples under certain types of transformations that preserve the class labels. Data generated under such label-preserving transformations will ameliorate the prediction invariance and generalization capacity pertaining to the neural networks. Data augmentation using label-preserving transformations proved to be effective for neural network training in terms of making invariant predictions.
The variation in handwriting style represents a potential problem with regards to online character. To resolve such problem, the concept of character distortion is introduced in order to generate numerous training samples. In fact, character distortion is produced via applying an affine transformation and its variants to the training data. The distortion is applied by performing flipping and  rotation (changing the angle of orientation).
Data augmentation  maintained by different scripts emanating from different writers who do not the same style of writing.
\begin{figure}[thpb]
      \centering
      \includegraphics[height=4cm,width=8cm]{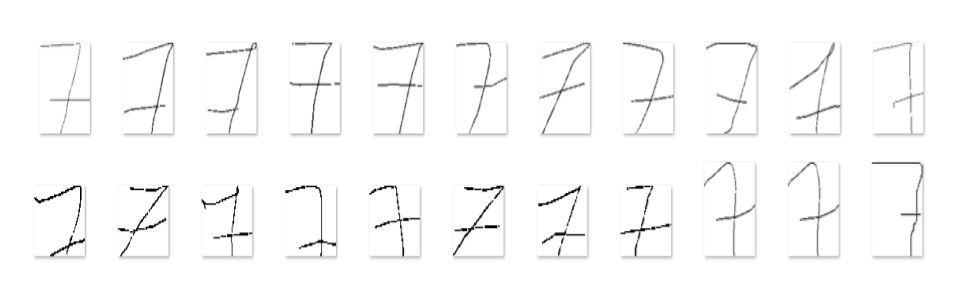}
      \caption{Distortion of digit 7.}      
\end{figure}  
\subsection{Training Network} 
To improve the performance of  two our proposed systems, various techniques are used that we will illustrate below.
\subsubsection{Framewise/Crisper Training}
We opt this technique of training with two proposed systems OnHRS-LSTM and OnHR-convLSTM which each segmented trace or not segmented trace are belong only to one class(one character,one word).
\begin{algorithm}
\caption{Crisper Training Algorithm} \label{alg:euclid}
\begin{algorithmic}
\Repeat
  \ForAll  
   \State {online script data (x, y, z) not segmented or segmented into basic perceptual codes: each data belongs to one class}         
    \State Use LSTM for training
   \EndFor
\Until{data=0}\\
 Use the trained network to predict the labels
\end{algorithmic}
\end{algorithm}
Each data belongs to one label. It means that Framewise method computes the forced alignments of the input  sequence with target output of the correct word/char sequence. We train the network to classify each sequence individually, based on cross-entropy cost function. \cite{Bluche2015}
Algorithm 1 describes our method of training, which consists in the fact that one sequence corresponds only to one predicted class.
\subsubsection{Training with Fuzzy Ground Truth}
To perform the network performance of our first proposed model, OnHRS-LSTM, each segmented data belongs to all classes with certain membership.
To define the value of membership, we calculate the segmented data which has the longest series of elliptic-strokes  $\{l_i\}$. Then, we calculate the Euclidian distance between the length  of class of reference and the actual data segmented  $\{l_r\}$ and $\{l_i\}$. 
\begin{algorithm}
\caption{Training with Fuzzy Ground Truth Algorithm} \label{alg:euclid}
\begin{algorithmic}
\Repeat 
   \ForAll \State {online script data (x, y, z) segmented into BE-parameters, define  $\{l_r\}$ }
   \State Euclidian distance between $\{l_r\}$ and $\{l_i\}$
   \State Use LSTM for training
   \EndFor
\Until{data=0}\\
 Use the trained network to predict the labels
\end{algorithmic}
\end{algorithm}
\begin{table}[h]
\caption{Common Characters Confusions in Latin/Arabic Script.}
\label{table_example}
\begin{center}

\begin{tabular}{ |c | c |c |c|c |c |c|}
\hline
\multicolumn{7}{|c|}{
\textbf{Latin Script}} \\
\hline
P & m & a & n & p & i & e\\ 
p & n & c & u & b & l & c \\
 \hline
\multicolumn{7}{|c|}{\textbf{Arabic Script}} \\
\hline\setcode{utf8}\<س > & \setcode{utf8}\<ع> &\setcode{utf8}\<د>  &\setcode{utf8}\<ص> & \setcode{utf8}\<ف> & \setcode{utf8}\<خ>& \setcode{utf8}\<ر>  \\ 
\setcode{utf8}\<شِ>  &\setcode{utf8}\<غ > & \setcode{utf8}\<ذ> & \setcode{utf8}\<ض>  & \setcode{utf8}\<ق>  & \setcode{utf8}\<ح> & \setcode{utf8}\<و> \\
\hline
\end{tabular}
\end{center}
\end{table}
\vspace{0pt}
\hspace{0pt}
\begin{table*}[h]
\caption{Comparison of recognition rate of System Perceptual Codes-LSTM with other systems.}
\label{table_example}
\begin{tabular}{|c|c|c|c|c|}
\hline
\rowcolor{Gray}
System	& Database &	Methods &	RR Framewise & RR Fuzzy Ground Truth\\
\hline
 Tagougui2014	& LMCA+ manually segmented characters form ADAB  &	DNN & 82.12 $\%$  & - \\
\hline
 Tagougui2014 &	LMCA+ manually Segmented characters form ADAB & MLP NN /HMM system 	& 96.45$\%$& -\\
\hline
OnHSR-LSTM	& LMCA+ manually segmented
characters form ADAB  &	BCP+LSTM & 91.50$\%$ & 97.50$\%$\\
\hline
OnHSR-LSTM &	MAYASTROUN & BCP+LSTM	& 94.50
$\%$ &  98.5 $\%$\\
\hline
OnHSR-LSTM	& ADAB
 & BCP+LSTM & 90.00 $\%$ & 96.00 $\%$\\
\hline
\end{tabular}
\end{table*}
\section{Single Character Experiments}
The pipeline of our system 1 , OnHSR-LSTM,  consists in obtaining the online script, segmenting it into  strokes, and classifying them into EPCs. 
After acquiring EPCs, we apply LSTM in order to obtain BPCs, and then eventually we apply LSTM in order to identify the character or shape or word corresponding of the initial script. The evaluation of the performance pertaining to our two proposed systems, OnHSR-LSTM $\&$ OnHR-convLSTM, relies to 5 multilingual databases.
\subsection{Arabic Database}
Factors do not generalize very well to a one , thus making the task of recognition difficult. Although Arabic is ranked  the fourth most spoken language in the world after Chinese, English and Spanish, to the best of our knowledge, there is  works in the field of Arabic text recognition, which pave the way for exploration of such favorable area of research. Arabic script, written from right to left, is a cursive script.
Arabic script, written from right to left is a cursive script in both printed and handwritten versions. Comparing to Latin scripts, it involves of strokes and dots above and under letters, along with, the connectivity of characters. There are 28 letters and the shape of those letters change depending on their position in the word: preceded  and/or  followed by other letters or isolated. Arabic letters can take four different shapes depending on their position in the word (first, middle, last and isolated) as illustrated in the Figure 1.  Some letters can take different forms, for example the letter Ghayen \setcode{utf8}\<(ـغ, غ, ـغـ, غـ)>.
We find that some Arabic letters sets share the same shapes and only a dot makes a difference between them such as or \setcode{utf8}\<(س , ش)> or
\setcode{utf8}\<(ح , خ  , ج)>. Arabic writing recognition, it is a hard task since numerous letters posses different numbers of positions and dots. In most cases, Arabic writing is devoid of vowels. The meaning of the word is frequently determined by the context of the sentence. Thus, all vowels are not considered in our work.
However, Latin script, is  considered as the easiest script, it serves the largest family of a total of 59 languages \cite{keysers2017}. It composes of 26 characters.
Our work based on the theory of perception i.e.: all we perceive is a form including handwriting scripts. Handwriting Recognition is feasible, even if their constitutions is missing based on the RBC approach\cite{Sourour2008,Sourour2012}. Handwriting is a visual scene perceived in order to decode the contained message as any visual scene presenting illusions. The existence of perceptual illusions in handwriting makes its recognition, identification and understanding difficult.  The main assumption of the proposed theory is that handwriting involves a concatenation of perceptual codes grouped together so as to obtain a shape, a character or a digit. Hence, the perceptual segmentation2 makes recognition easier by perceiving a character as a group of BPC.
\begin{figure}[thpb]
    \centering
      \includegraphics[height=6cm,width=6cm]{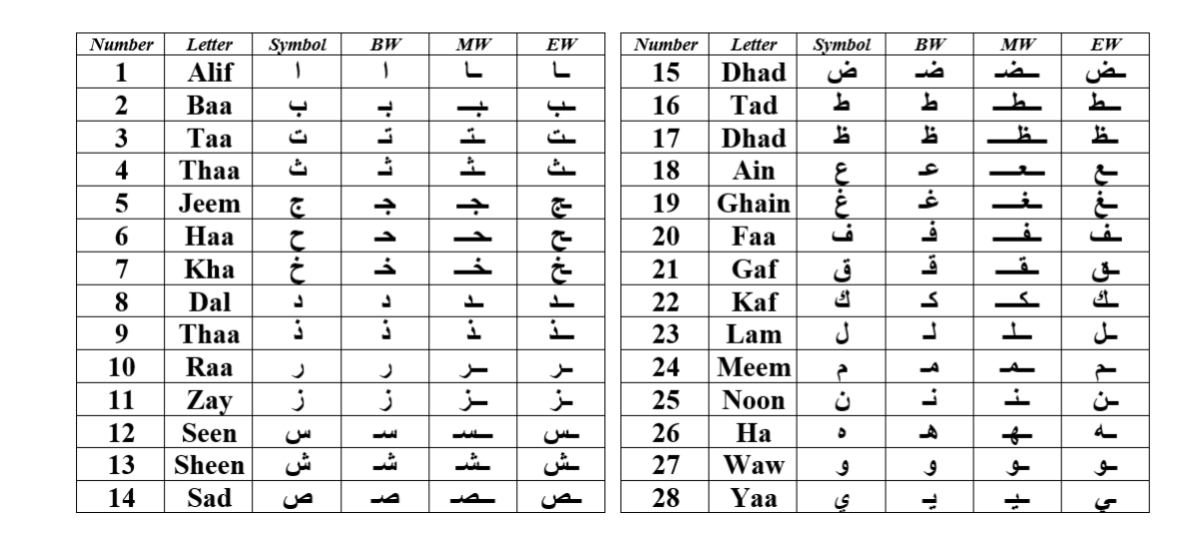}
   \caption{Arabic Alphabets and their forms at different positions.}
\end{figure}
In order to compare our proposed works, we need to use the same lexicon which is Arabic in  this section, extracted from MAYSTROUN \cite{Sourour2012} , LMCA and manually segmented from ADAB Database. 
MAYSTROUN, is composed of  400 letters, 300 digits, 200 words and 200 Arabic texts written by 15 writers\cite{Njah44}.
Kherallah et al. \cite{Kharallah2002} used Circular trajectory modeling and NN for Arabic digits recognition, wherein each digit is repeated 1000 times, and they obtained 95\% as recognition rate. Compared to the results of [12], we obtained better recognition by achieving 98.5\% of Arabic digit recognition. In Table 2 we compare our results with the results achieved by Tagougui et al.\cite{Tagougui14},\cite{Najiba2017}. In fact, with reference to Table 2, LMCA and manually segmented from ADAB Database display noticeable performances as compared to other already developed methods. Indeed, LMCA (Lettres Mots Chiffres Arabes) database\cite{Njah2011,Njah2012} contains 30.000 shapes for ten digits, 100.000 shapes for 56 Arabic letters and 500 Arabic words. 55 respondents were hosted to participate in the development of the handwritten LMCA. We used 54 shapes instead of 56 Shapes. Some letters are manually segmented Letters and ligatures,  extracted from the four sets (a,b,c and d) of ADAB database \cite{Kherallah2008} to recognize isolated online Arabic handwritten characters.
To test our system OnHSR-LSTM, we use 24861 samples for training and test, such as digits and Arabic letters (beginning, middle and end of the word) incorporating various styles of handwriting.
The over-fitting can be  reduced considerably by randomly omitting half of the hidden units on each training case by using Dropout. Random dropout yields remarkable improvements to our proposed. We perform our system 1  with Fuzzy Ground Truth in the range 94$\%$to 98.2$\%$ as recognition rate. We attempt to use different combinations of one BPC or character in order to prevent the distortion of the model during the training step.
Table 6 reveals that OnHSR-LSTM, describes both static and dynamic aspects of handwriting and by means of perceptual codes we codify the scripts into sequence of codes, in the form of an equation.\label{h1,h2}
Moreover, it demonstrates that the method of training fuzzy Ground Truth improved the recognition rate RR, yielding competitive results compared to  similar works.
Then, those results indicate the fact that the results increase intensively by applying data augmentation and dropout mechanisms.
As shown in Table 7, our Perceptual Codes-LSTM system achieves the best recognition rate, especially when we use CTC and Fuzzy Ground Truth methods during training.
Table 7 boosts the fact that Fuzzy LSTM is endowed with a more efficient performance,compared to framewise proposed approaches in terms of recognition rate.
\subsection{Latin Database}
The problem with comparing online handwritten character recognition systems lies to the few number of available databases.
Accordingly, we find an old database UNIPEN \cite{Guyon1994}  which is used frequently, however,its evaluation is restricted to single Latin character recognition. It is used in \cite{keysers2017}. Thus, we attempt to compare it with  our recognizer system 1. The results are summarized in Table 8. We observe that the recognition rate is very competitive, especially with set 1(c), due to CTC and Fuzzy Ground Truth as well as for the prepossessing applying in online script.  \\
\begin{table*}[h]
\caption{Comparison of recognition rate with Google system with UNIPEN set 1(c).}
\label{table_example}
\begin{center}
\begin{tabular}{|c|c|c|c|}
\hline
\rowcolor{Gray}
System	&	Methods &	RR with Framewise & RR with Fuzzy Ground Truth \\
\hline
 \cite{keysers2017}  &	Language model & 94.9 $\%$ & -\\
\hline
OnHSR-LSTM & BCP+LSTM 	& 90.50 $\%$ & 97.5$\%$\\  
\hline
\end{tabular}
\end{center}
\end{table*}
So far, based on the aforementioned two precedent tables, using fuzzy training gives rise to an enhancement in terms of classification accuracy. In order to  further ensure the strength of the proposed approaches, additional tests are performed on another database IRONOFF\cite{article1999}. The testing results of recognition of digits are illustrated in Table 8.
In fact, this table demonstrates that our system 2, OnHR-convLSTM, achieves efficient results compared to our system 1, OnHSR-LSTM.
\begin{table}[h]
\caption{Comparison of recognition rate with IRONOFF Database.}
\label{table_example}
\begin{center}
\begin{tabular}{|c|c|c|}
\hline
\rowcolor{Gray}
System	&	Methods &	Recognition Rate \\
\hline
OnHSR-LSTM & BCP+LSTM
	& 97.5$\%$\\
\hline

OnHR-convLSTM & convLSTM
	& 98.00$\%$\\
\hline
\end{tabular}
\end{center}
\end{table}
To allow a fair comparison between framewise and fuzzy training, the performances of the two proposed recognizers is shown in the next figure which describes the evolution of predicted and test fuzzy ground truth,illustrating that the two curves are close to each other.\\

\begin{figure}[thpb]
      \centering
      \includegraphics[height=4cm,width=8cm]{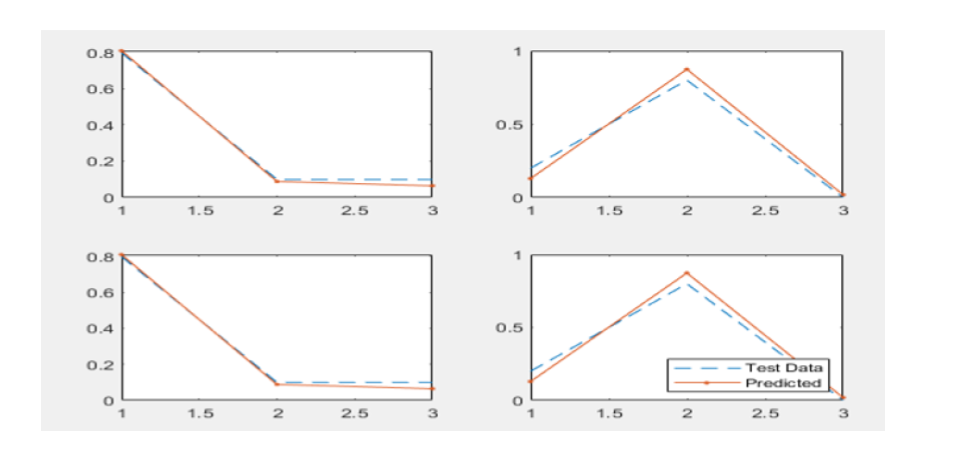}
      \caption{Classes Predicted and Correct ones.}      
\end{figure} 
\subsection{Our Mobile Database}
Based on mobile application of [7], we created a mobile database for acquired data from different persons with different styles of writing in order to examine the performance of our system.
The (x, y, z) coordinates are stored in text file of online written script. \\
The online script can be presented as a variable length sequence.
\begin{equation}
input = [({x_1},{y_1},{z_1}),({x_2},{y_2},{z_3}),...,({x_n},{y_n},{z_n})]\\
\end{equation}
where $\{x_i\}$ and  $\{y_i\}$ are the xy-coordinates of the pen movements and $\{z_i\}$ is 0 when the pen is up and is 1 when the pen is down.\\ 
\begin{figure*}[!ht]
      \centering
      \includegraphics[width=\textwidth]{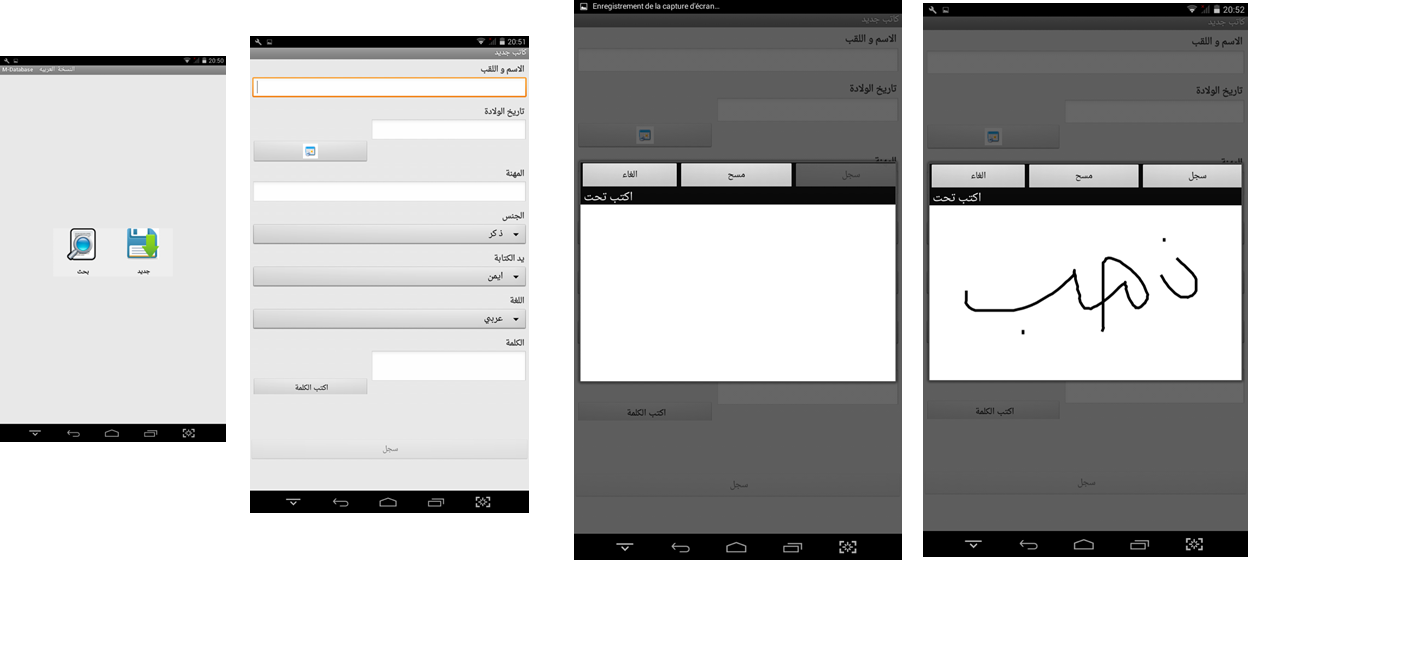}
      \caption{Interfaces of Mobile Database.}
\end{figure*} 
Moreover, experiments  were conducted on this mobile database. Correspondingly,  we remark that the results are very competitive with reference to previously published results, which presented in Table 8 \cite{Hanen2017} \cite{Akouaydi2016}.
\begin{table}[h]
\caption{Comparison of recognition rate with previous results system with Mobile Database .}
\label{table_example}
\begin{center}
\begin{tabular}{|c|c|c|}
\hline
\rowcolor{Gray}
System	&	Methods &	Recognition Rate \\
\hline
OnHR-convLSTM  & ConvLSTM & 96.50$\%$\\
\hline
OnHSR-LSTM & BCP+LSTM	& 98.40$\%$\\
\hline
\end{tabular}
\end{center}
\end{table}
\section{Word Recognition Experiments}
\subsection{Word Recognition Experiments on ADAB}
ADAB is a database of  Arabic online handwritten words\cite{adab}. To evaluate our two systems we used it and  good results are achieved as describes table 10 below.Table 10 shows that our system 2, OnHR-convLSTM, achieves best results compared to our system 1,OnHSR-LSTM .

\begin{table}[h]
\caption{Comparison of recognition rate with ADAB Database.}
\label{table_example}
\begin{center}
\begin{tabular}{|c|c|c|}
\hline
\rowcolor{Gray}
System	&	Methods &	Recognition Rate \\
\hline
OnHSR-LSTM & BCP+LSTM
	&96.50$\%$\\
\hline
OnHR-convLSTM & ConvLSTM
	& 97.0$\%$\\
\hline
\end{tabular}
\end{center}
\end{table}
\subsection{Word Recognition Experiments on IRONOFF}
The IRONOFF database\cite{article1999} contains   31346 isolated word
 from a 196 word lexicon (French and English). Although the database contains both
online and offlifne information of the handwriting signals, only the online information is
used for our experiments. Table 9 reports the performance of our recognition systems 
Again, we achieve a very competitive rate with respect to other published results we are aware of.
\begin{table}[h]
\caption{Comparison of recognition rate with IRONOFF Database.}
\label{table_example}
\begin{center}
\begin{tabular}{|c|c|c|}
\hline
\rowcolor{Gray}
System	&	Methods &	Recognition Rate \\
\hline
OnHSR-LSTM & BCP+LSTM
 & 93 $\%$\\
\hline
OnHR-convLSTM & ConvLSTM
	& 98$\%$\\
\hline
\end{tabular}
\end{center}
\end{table}
\section{Noise}
In order to study the robustness of the proposed recognizer system, a randomly noise signal is applied  in both training and testing databases.  Noise in handwriting can be ascribed to the support of writing, the person who write,  and Parkinson person who produces trembled script.
\begin{table}[h]
\caption{Recognition Rate with LMCA and manually segmented from ADAB Database with noise .}
\label{table_example}
\begin{center}
\begin{tabular}{|c|c|c|c|}
\hline
\rowcolor{Gray}
 System	&	Methods &	Recognition Rate \\
\hline
OnHSR-LSTM & BCP+LSTM & 96.00  $\%$\\
\hline
OnHR-convLSTM & (x,y,z)+ConvLSTM & 50.45$\%$\\
\hline
OnHR-convLSTM & (x,y,z)+ConvLSTM +teta+4 EPCs
	& 74.41$\%$\\
\hline
\end{tabular}
\end{center}
\end{table}
Figure 15 illustrates how our Beta-elliptic segmentation method eliminates trembling from script. Table 10 demonstrates that  LSTM with Beta-elliptic features yield best accuracy compared to (x,y,z),  only because the Beta-elliptic method can  improve the online script and eliminate, displaying its robustness with respect to noise. Therefore, our perceptual method, based on its preprocessing step can improve the quality of online script via removing noise and applying sampling, which is not available with OnHR-convLSTM system  that neglects the preprocessing and improving step of the signal of the input handwritten script. However, OnHR-convLSTM with teta+4 EPCs provides result compared to only (x,y,z). Thus, the elliptic model is not only a segmentation model but is also with its methods of preprocessing, capable of improving the quality of acquired script, thus yielding competitive recognition rate.
\begin{figure}[thpb]
      \centering
      \includegraphics[height=4cm,width=8cm]{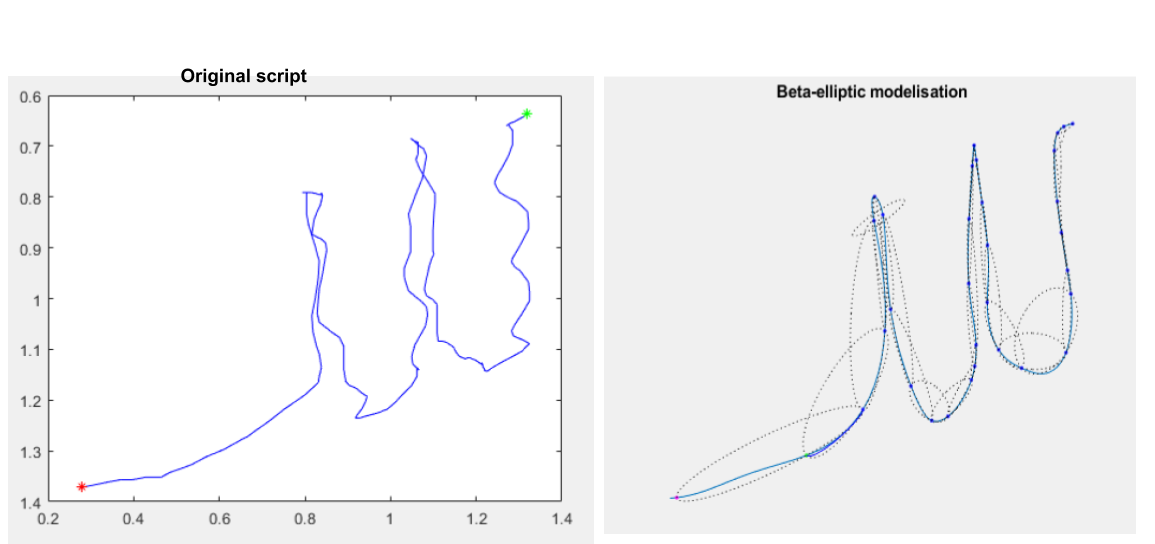}
      \caption{Sample of input noise $\&$ output}      
\end{figure} 
Beta-elliptic model can eliminate trembling from script. But also can detect with its parameters velocity and produce a different shape then a normal one
So, Beta-elliptic a good solution to detect some assumptions of trembling in handwriting and can assists in the decision process leading to the diagnosis of some diseases related to trembles of handwriting or movements.
Figure 16 illustrate the difference beta-elliptic points between normal script and trembled script.
\begin{figure}[thpb]
      \centering
      \includegraphics[height=4cm,width=8cm]{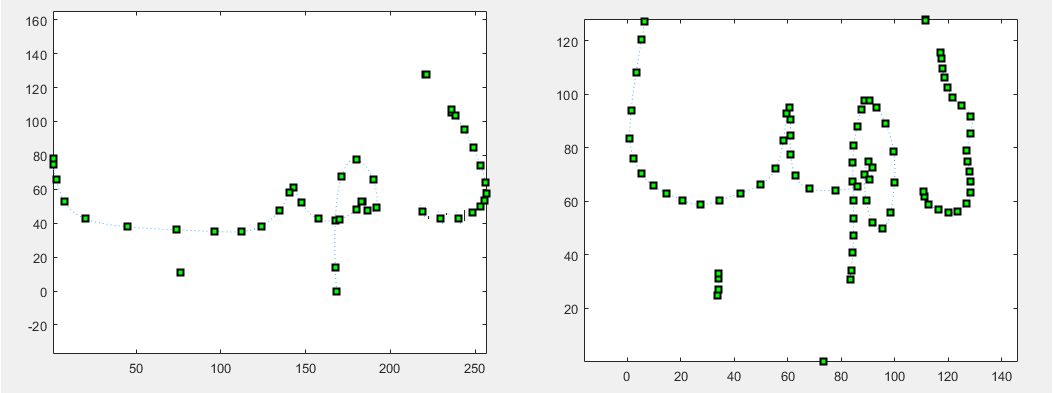}
      \caption{Sample of input noise $\&$ output word dhahab \setcode{utf8}\<ذهب> }      
\end{figure} 

\begin{figure}
      \centering
      \includegraphics[height=4cm,width=8cm]{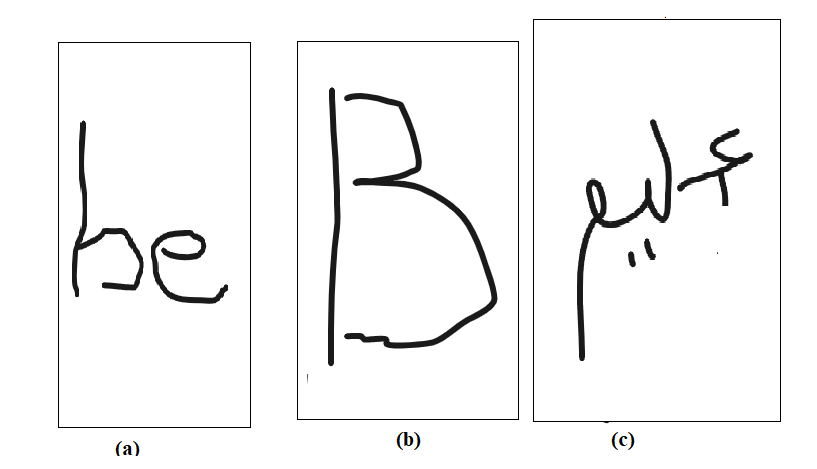}
      \caption{Handwriting perceptual illusions : (a): the word “he” or “be”,
(b): the letter “B” or the number “13”, (e): the two Arabic words “ aalimon: \setcode{utf8}\<عليم>
 ”  or “ alimon : \setcode{utf8}\<اليم>”.}
\end{figure}
\
\section{Perceptual Errors}
The existence of perceptual illusions in handwriting makes its recognition, identification and understanding difficult. Figure 16 presents some examples of handwriting perceptual illusions.
Those errors can be performed by our recognition language system.
In Figure 16.(a) what do you see? : This handwriting shape can be identified as a letter "O" or a digit "0", according to the context it appears in.
The presented handwriting perceptual illusions proved that finding evidence, whether or not humans go through an identification process as a part of perception, has been difficult.

\section{Conclusion}
Two systems for online handwriting recognition are presented in this paper. The first is  based on the fact that handwriting is composed of a group of Perceptual Codes and the second is generative system based on convolutional LSTM.  We validate our system with three databases and mobile database, containing digits, letters and words in Arabic language. Good results are obtained almost identical to those produced by the human perceptual system. Encouraged obtained  recognition rate is equal to 98.5\%. We plan to build and test our proposed systems recognition in a large-scale database and other scripts.The perspective of improvement of our method is located in order to reach the stage of a sentence. Thanks to beta-elliptic model for handwriting strokes generation, we obtain supplementary information about dynamic domain. Furthermore, through this segmentation model, we can improve the quality of acquired trace.

%

\ifCLASSOPTIONcompsoc
  \section*{Acknowledgments}
\else
  \section*{Acknowledgment}
\fi
The research leading to these results received funding from the Ministry of Higher Education and Scientific Research of Tunisia under the grant agreement number LR11ES48.
Moreover, this project was carried out under the MOBIDOC scheme, funded by the EU through the EMORI program and managed by the ANPR. 

\ifCLASSOPTIONcaptionsoff
  \newpage
\fi



%
\bibliographystyle{IEEEtran}
\bibliography{mybib}

%

\begin{IEEEbiography}[{\includegraphics[width=1in,height=1.25in,clip,keepaspectratio]{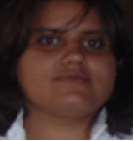}}]{Hanen Akouaydi}
currently a PhD student in computer science at the National School of Engineers of Sfax. She obtained her engineering degree in Computer Science from Higher Institute of Multimedia Arts of Manouba in 2012. Her research interests include online handwriting recognition and deep learning. She is a member of the IEEE Student and the IEEE Computer Society and affiliate to REGIM Laboratory.
\end{IEEEbiography}
\begin{IEEEbiography}[{\includegraphics[width=1in,height=1.25in,clip,keepaspectratio]{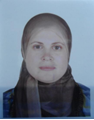}}]{Sourour Njah} currently an Assistant Professor at the High Institute of Commerce of Sfax. She earned the M.S. degree in Computer Science from Faculty of Economics and Management of Sfax in 2003 and the Ph.D. degree in Computer Engineering from the National School of Engineers of Sfax in 2013. Her research domains are cursive handwriting analysis and fuzzy logic. She is a member of the IEEE and affiliate to REGIM Laboratory.
\end{IEEEbiography}
\begin{IEEEbiography}[{\includegraphics[width=1in,height=1.25in,clip,keepaspectratio]{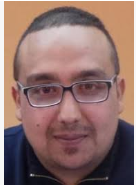}}]{Wael Ouarda}
is currently a Postdoctoral Researcher at the National School of Engineers of Sfax, Tunisia. He received earned the his M.S.
degree in Computer Science from the INSA Lyon, France in 2010 and the Ph.D. degree in Computer Engineering from the National
School of Engineers of Sfax in 2016. His research focused on Bio- metric systems, Information Fusion and Pattern Recognition. He
is a member of the IEEE and he is affiliate to REGIM Laboratory.
\end{IEEEbiography}
\begin{IEEEbiography}[{\includegraphics[width=1in,height=1.25in,clip,keepaspectratio]{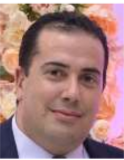}}]{Anis Samet} the director of the site SIFAST in Tunisia since 2013. He received his M.S. in IT applied to management from the Faculty of Management of Sfax in 2004. His work focused on Team management, E-learning, Quality and process management,Business Intelligence,Piloting of teams,Business Unit, site, Piloting  projects, etc.
\end{IEEEbiography}

\begin{IEEEbiography}[{\includegraphics[width=1in,height=1.25in,clip,keepaspectratio]{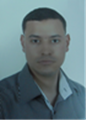}}]{Thameur Dhieb}
presently a PhD student of computer science at the Higher Institute of Computer Sciences and Communication Techniques of Hammam Sousse. He obtained his M.S. degree in Computer Science from the Faculty of Economics and Management of Sfax in 2012. His research interests include writer identifica-tion and signature verification. He is a member of the IEEE Student and the IEEE Computer Society and affiliate to REGIM Laboratory.
\end{IEEEbiography}

\begin{IEEEbiography}[{\includegraphics[width=1in,height=1.25in,clip,keepaspectratio]{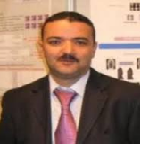}}]{Mourad Zaied}
actually a Professor at the National School of Engineers of Gabes. He earned the Ph.D. degree and then the HDR both in Electrical and Computer Engineering in 2013. He is IEEE Senior member since 2000.
He received his HDR, and his Ph.D degrees in Computer Engineering and the Master of science from the National Engineering School of Sfax respectively in 2013,2008 and in 2003. He obtained the degree of Computer Engineer from the National Engineering School of Monastir in 1995.  Since 1997 he served in several institutes and faculties in  the university of Gabes as teaching assistant. His research interests include Computer Vision and Image and video analysis. These research activities are centered on Wavelets and Wavelet networks and their applications to data classification and approximation, pattern recognition and image, audio and video coding and indexing.
\end{IEEEbiography}
\begin{IEEEbiography}[{\includegraphics[width=1in,height=1.25in,clip,keepaspectratio]{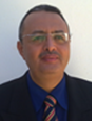}}]{Adel M. Alimi} currently a Professor at the National School of Engineers of Sfax. He obtained his Ph.D. degree and then his HDR both in Electrical and Computer Engineering in 1995 and 2000 respectively. His research interest includes applications of intelligent methods and vision systems. He is a guest editor of several special issues of international journals like Soft Computing. He is IEEE Senior member since 2000 and he is the director of REGIM Laboratory.
\end{IEEEbiography}
\vfill



\end{document}